\documentclass{www2009-accepted}
\pdfoutput=1
\usepackage{xspace}
\usepackage{graphicx}
\usepackage[overload]{textcase}
\usepackage{times}
\usepackage{amssymb}
\usepackage{color}
\definecolor{light-gray}{gray}{0.55}
\usepackage{slashbox}
\usepackage{colortbl}
\usepackage{url}

\newcommand{\headerformat}[1]{\MakeTextUppercase{#1}} 

\newcommand{\unimodalinformal}{single-peaked\xspace}

\newcommand{\helpfulness}{helpfulness\xspace}
\newcommand{\Helpfulness}{Helpfulness\xspace}
\newcommand{\hvote}{\helpfulness vote\xspace}
\newcommand{\hvotes}{{\hvote}s\xspace}
\newcommand{\Hvotes}{\Helpfulness Votes}
\newcommand{\smark}{star rating\xspace}
\newcommand{\product}{product\xspace}
\newcommand{\products}{{\product}s\xspace}
\newcommand{\avgmarkcomp}{computed \product-average {\smark}\xspace}
\newcommand{\avgmarkdisplayed}{Amazon-displayed \product-average {\smark}\xspace}

\newcommand{\avgmarkcompabbrev}{computed star average\xspace}

\newcommand{\avgmarkdisplayedabbrev}{displayed star average\xspace}

\newcommand{\deviation}{deviation\xspace}
\newcommand{\deviations}{deviations\xspace}
\newcommand{\Deviation}{Deviation\xspace}
\newcommand{\hratio}{helpfulness ratio\xspace}
\newcommand{\Hratio}{Helpfulness ratio\xspace}
\newcommand{\hratios}{{\hratio}s\xspace}
\newcommand{\smarks}{{\smark}s\xspace}
\newcommand{\hevaluator}{helpfulness evaluator\xspace}
\newcommand{\hevaluators}{{\hevaluator}s\xspace}
\newcommand{\sdeviation}{signed {\deviation}\xspace}
\newcommand{\sdeviations}{signed {\deviations}\xspace}
\newcommand{\Sdeviations}{Signed {\deviations}\xspace}
\newcommand{\variance}{variance\xspace}

\providecommand{\hypothesis}{hypothesis\xspace}
\providecommand{\hypotheses}{hypotheses\xspace}
\providecommand{\conformity}{conformity\xspace}
\providecommand{\conformityhyp}{\conformity \hypothesis}
\providecommand{\bias}{individual bias\xspace}
\providecommand{\biases}{individual biases\xspace}
\providecommand{\biasprefix}{individual-bias\xspace}
\providecommand{\biashyp}{\biasprefix \hypothesis}
\providecommand{\brilliantprefix}{brilliant-but-cruel\xspace}
\providecommand{\brillianthyp}{\brilliantprefix \hypothesis}
\providecommand{\contenthypprefix}{quality-only\xspace}
\providecommand{\contenthyp}{\contenthypprefix \hypothesis}

\newcommand{\plagiarized}{``plagiarized''\xspace}
\newcommand{\plagiarism}{``plagiarism''\xspace}

\newcommand{\Plagiarized}{``Plagiarized''\xspace}

\newcommand{\textualquality}{text quality\xspace}

\newcommand{\moreh}{$\succ$}
\newcommand{\lessh}{$\prec$}
\newcommand{\morehu}{~}
\newcommand{\lesshu}{\morehu}
\newcommand{\ec}{\cellcolor[gray]{0.8} ~} 

\newcommand{\significantly}{significantly\xspace}
\newcommand{\significant}{significant\xspace}

\newcommand{\procofthe}{Proc.\xspace}
\newcommand{\journalof}{J.\xspace} 
\newcommand{\procof}{\procofthe}

\title{How Opinions 
are Received by Online Communities:\\
A Case Study on Amazon.com \Hvotes}

\usepackage[numbers,sort&compress]{natbib}\providecommand{\newblock}{}

 \newcommand{\lastname}[1]{#1}
   \newcommand{\dnm}{Danescu-Niculescu-Mizil}

\newcommand{\deptaffil}[1]{#1} 
\newcommand{\cucs}{\deptaffil{\affaddr{Dept. of Computer Science}\\}\affaddr{Cornell University}\\\affaddr{Ithaca, NY 14853 USA}}

   \numberofauthors{3} 
   \author{
       \alignauthor \hspace*{-.3in}\mbox{Cristian \lastname{\dnm}}\\
\cucs
\\\email{cristian@cs.cornell.edu}
       \alignauthor\hspace*{.16in}\mbox{Gueorgi \lastname{Kossinets}}\\ \affaddr{\hspace*{.19in} Google Inc.}\\  \deptaffil{\affaddr{\hspace*{.08in}\mbox{Mountain View, CA 94043 USA}}\\}  \email{\hspace*{.15in}gkossinets@google.com}
	  \and 
          \alignauthor Jon \lastname{Kleinberg}\\ \cucs \\\email{kleinber@cs.cornell.edu}
          \and \alignauthor Lillian \lastname{Lee}\\ \cucs \\\email{llee@cs.cornell.edu}
}

\begin{document}

\maketitle

\begin{abstract}
There are many on-line settings in which users publicly express opinions.  A
number of these offer mechanisms for other users to {\em evaluate} these
opinions; a canonical example is Amazon.com, where reviews come with
annotations like ``26 of 32 people found the following review helpful.''
Opinion evaluation appears in many off-line settings as well, including
market research and political campaigns.  Reasoning about the evaluation of
an opinion is fundamentally different from reasoning about the opinion
itself: rather than asking, ``What did Y think of X?'', we are
asking, ``What did Z think of Y's opinion of X?''
Here we develop a framework for analyzing and modeling opinion
evaluation, using a large-scale collection of Amazon book reviews as a
dataset.  We find that the perceived helpfulness of a review depends
not just on its content but also 
but also in subtle ways on how the expressed evaluation relates to other
evaluations of the same product. 
As part of our
approach, we develop novel methods
that take advantage of the phenomenon of review \plagiarism
to
control for the effects of text in opinion evaluation,
and we provide a simple and natural mathematical
model consistent with our findings.
Our analysis also allows us to distinguish among
the predictions of competing theories from sociology and
social psychology, and to discover unexpected differences 
in the collective 
opinion-evaluation behavior of user populations from
different countries.
\end{abstract}

\vspace{1mm}
\noindent
{\bf Categories and Subject Descriptors:}
H.2.8 [{\bf Database Management}]: {Database Applications -- Data Mining}

\vspace{1mm}
\noindent
{\bf General Terms:} Measurement, Theory

\vspace{1mm}

\newcommand{\xhdr}[1]{\paragraph*{{\bf #1}}}
\def\metadata{{metadata}}

\section{Introduction}

Understanding how people's opinions are received and
evaluated is a fundamental problem that arises in many
domains, such as
in marketing studies of the impact
of reviews on product 
sales, or in political science models
of how support for a candidate depends on the views he or
she expresses on different topics.  
This issue is 
 also
increasingly important in the 
user interaction dynamics
of large participatory Web sites.

Here we develop a framework for
 understanding and modeling
how opinions are evaluated within on-line communities.  The
problem is related to the lines of computer-science research
on opinion, sentiment, and subjective content 
\citep{Pang+Lee:08a}, but with a
crucial twist in its formulation that makes it fundamentally
distinct from
that body of work.  Rather than asking
questions of the 
form ``What did Y think of X?'', we are
asking, ``What did Z think of Y's opinion of X?''
Crucially, there are now three entities in the process
rather than two.  Such three-level concerns are widespread
in everyday life, and integral to any study of opinion
dynamics in a community.  For example, political polls will
more typically ask, ``How do you feel about Barack Obama's
position on taxes?''  
than ``How do you feel about
taxes?'' or ``What is Barack Obama's position on taxes?''
(though all of these are useful questions in different
contexts). 
Also, Heider's theory of {\em structural balance} in social psychology
seeks to understand subjective relationships 
by considering sets of three entities at a
time as the basic unit of analysis. 
But there has been
relatively little investigation of how these three-way
effects shape the dynamics of on-line interaction, and this
is the topic we consider here.

\vspace*{-.1in}

\xhdr{The Helpfulness of Reviews} The evaluation of opinions
takes place at very large scales every day at a number of
widely-used Web sites.  Perhaps most prominently it is
exemplified by one of the largest online e-commerce
providers, Amazon.com, whose website includes not just
product reviews contributed by users, but also evaluations
of the helpfulness of these reviews.  
(These consist of annotations that say things like,
``26 of 32 people found the following review helpful'', with
the corresponding data-gathering question, ``Was this review
helpful to you?'')  Note that each review on Amazon thus
comes with both a
{\em \smark}
--- the number of
number of stars it assigns to the product --- and a
{\em 
  \hvote } --- the 
information that $a$ out of $b$ people
found the review
itself
 helpful.
(See Figure \ref{fig:plagiarism_example} for two examples.)
This distinction reflects precisely the kind of opinion
evaluation we are 
considering: in addition to the
question ``what do you think of book X?'', users are also
being asked ``what do you think of user Y's review of book
X?''  A large-scale snapshot of Amazon reviews and \hvotes
will form the central dataset in our study, as detailed
below.

The factors affecting human \helpfulness evaluations are not
well understood.  There has been a small amount of work on
automatic determination of \helpfulness, treating it as a
classification or regression problem with Amazon \hvotes
providing labeled data
\citep{Kim+al:06a,Ghose+Ipeirotis:07a,Liu+al:07a}.  
Some of this
research has indicated that the 
\hvotes of reviews 
are not necessarily strongly correlated with  certain measures of review quality;
for example, Liu et
al. found that when they provided independent human
annotators with Amazon review text and a precise
specification of helpfulness in terms of the thoroughness of
the review, the annotators' evaluations differed
significantly from the \hvotes observed on Amazon.

All of this suggests that there is in fact a subtle relationship
between two different meanings of ``helpfulness'':
helpfulness in the narrow sense --- does this review help
you in making a purchase decision? ---
and helpfulness ``in the wild,'' as defined by the way in which
Amazon users evaluate each others' reviews in practice.
It is a kind of dichotomy familiar from the design of
participatory Web sites, in which a presumed design
goal --- that of highlighting reviews that are helpful 
in the purchase process --- becomes intertwined with
complex social feedback mechanisms.
If we want to understand how these definitions interact
with each other, so as to assist users in interpreting 
helpfulness evaluations, we need to elucidate what
these feedback mechanisms are and how they affect the
observed outcomes.

\xhdr{The present work: Social mechanisms underlying
  helpfulness evaluation} In this paper, we formulate and
assess a set of theories that govern the evaluation of
opinions, and apply these to a dataset consisting of over
four million reviews of roughly 675,000 books on Amazon's
U.S. site, as well as smaller but comparably-sized corpora
from Amazon's U.K., Germany, and Japan sites.
The
resulting analysis provides a way to distinguish among
competing hypotheses for the social feedback mechanisms at
work in the evaluation of Amazon reviews: we offer evidence
against certain of these mechanisms, and show how a simple
model can directly account for a relatively complex
dependence of helpfulness on review and group
characteristics.  We also use a novel experimental
methodology 
that takes advantage of the phenomenon of review \plagiarism 
to control for the text content of the reviews,
enabling us to focus 
exclusively
 on factors outside the
text that affect 
\helpfulness 
evaluation.

In our initial exploration of non-textual factors 
that are correlated with helpfulness evaluation on Amazon,
we found a broad collection of effects at varying levels of 
strength.\footnote{For example, on the U.S. Amazon site, 
we find that reviews from authors
with addresses in U.S. territories outside
the 50 states get consistently lower \hvotes.
This is a persistent effect whose possible bases lie outside
the scope of the present paper, but it illustrates the ways in which
non-textual factors can be correlated with helpfulness evaluations.
Previous work has also noted that {\em longer} reviews 
tend to be viewed as more helpful; ultimately it is a definitional
question whether review length is a textual or non-textual
feature of the review.}
A significant and particularly wide-ranging set of
effects is based on the relationship of 
a review's 
\smark
to the 
\smarks
of other reviews for the same product.
We view these as fundamentally {\em social} effects,
given that they are based on the relationship of one user's
opinion to the opinions expressed by others in the same setting.
\footnote{A contrarian might put forth the following non-socially-based alternative hypothesis: 
the people evaluating review \helpfulness are
considering actual product quality rather than other reviews, but
aggregate opinion happens to coincide with objective product
quality.  This hypothesis is not consistent with our experimental
results. However, in future work it might be interesting to 
directly control for product quality.}

Research in the social sciences provides a range of well-studied
hypotheses for how social effects influence a group's
reaction to an opinion, and these provide a valuable starting
point for our analysis of the Amazon data.
In particular, 
we consider the following three broad classes of theories, as well as
a fourth straw-man hypothesis that must be taken into account.
\begin{itemize}
\item[(i)] {\em The \conformityhyp.}  One hypothesis, with
roots in the social psychology of conformity \cite{Bond+Smith:96a}, 
holds that a
review is evaluated as more helpful when its \smark is closer
to the consensus \smark for the product --- for example, when
the number of stars it assigns is close to the average number 
of stars over all reviews.
\item[(ii)] {\em The \biashyp.}  Alternately,
one could hypothesize that when a user considers a review, 
he or she will rate it more highly if it expresses an opinion that
he or she agrees with.\footnote{Such a principle is also
supported by structural balance considerations from social psychology;
due to the space limitations, we 
omit a discussion of this here.}
Note the contrasts and similarities with the previous hypothesis:
rather than evaluating whether a review is close to the mean opinion,
a user evaluates whether it is close to their own opinion.
At the same time, one might expect that if a diverse range of individuals
apply this rule, then the overall helpfulness evaluation could
be hard to distinguish from one based on conformity; 
this issue turns out to be crucial, and we explore it further below.
\item[(iii)] {\em The brilliant-but-cruel hypothesis.}
The name of this hypothesis comes from studies performed by
\citet{Amabile:83a} that support the argument that
``negative reviewers [are] perceived as more intelligent,
competent, and expert than positive reviewers.''
One can recognize everyday analogues of this phenomenon;
for example, in a research seminar, a dynamic may arise in which the
nastiest question is consistently viewed as the most insightful.
\item[(iv)] {\em The \contenthypprefix straw-man hypothesis.} 
Finally, there is a challenging methodological complication
in all these styles of analysis: without specific evidence,
one cannot dismiss out of hand 
the possibility that helpfulness is being evaluated purely
based on the textual content of the reviews, and that 
these non-textual factors are simply correlates of textual quality.
In other words, it could be that people who write long reviews,
people who assign particular \smarks in particular situations,
and people from Massachusetts all simply write reviews that
are textually more helpful --- and that users performing 
helpfulness evaluations are simply reacting to the text
in ways that are indirectly reflected in these other features.
Ruling out this hypothesis requires some means of controlling
for the text of reviews while allowing other features to vary,
a problem that we also address below.
\end{itemize}
We now consider how data on \smarks and \hvotes
can support or contradict these hypotheses, and what
it says about possible underlying social mechanisms.

\xhdr{Deviation from the mean} A natural first measure to
investigate is the relationship of a review's \smark to the
mean \smark of all reviews for the product; this, for
example, is the underpinning of the conformity hypothesis.
With this in mind, let us define the {\em \hratio} of a
review to be the fraction of evaluators who found it to be
helpful (in other words, it is the fraction $a/b$ when $a$
out of $b$ people found the review helpful), and let us
define the {\em product average} for a review of a given
\product to be the average \smark given by {\em all} reviews
of that \product.
We find (Figure~\ref{fig:unsigned-diff-novar})
that the median \hratio of reviews decreases monotonically
as a function the absolute difference between their \smark
and the product average.  (The same trend holds for other
quantiles.)  In fact the dependence is surprisingly smooth,
with even seemingly subtle changes in the differences from
the average having noticeable effects.

This finding on its own is consistent with the conformity hypothesis:
reviews in aggregate are deemed more helpful when they are close
to the product average.
However, a closer look at the data raises complications,
as we now see.
First, to assess the brilliant-but-cruel hypothesis, 
it is natural to look not at the absolute difference between
a review's \smark and its product average, but at
the {\em signed difference}, which is positive or negative
depending on whether the \smark is above or below the average.
Here we find something a bit surprising (Figure~\ref{fig:signed-diff-novar}).
Not only does the median helpfulness as a function of signed
difference fall away on both sides of $0$; it does so 
{\em asymmetrically}: slightly negative reviews are punished
more strongly, with respect to \helpfulness evaluation,
than slightly positive reviews. 
In addition to being at odds with the brilliant-but-cruel
hypothesis for Amazon reviews, this observation poses problems for
the conformity hypothesis in its pure form.
It is not simply that closeness to the average is rewarded;
among reviews that are slightly away from the mean, 
there is a bias toward overly positive ones.

\xhdr{Variance and individual bias}
One could, of course, amend the conformity hypothesis so
that it becomes a ``conformity with a tendency toward positivity'' hypothesis.
But this would beg the question; it wouldn't suggest any
underlying mechanism for where the favorable evaluation of positive
reviews is coming from.
Instead, to look for such a mechanism, we consider versions of the
individual-bias hypothesis.
Now, recall that it can be difficult to distinguish conformity
effects from individual-bias effects in a domain such as ours:
if people's opinions
 (i.e., \smarks) 
for a product come from a \unimodalinformal distribution
with a maximum near the average, then the composite of their
individual biases can produce overall \hvotes
that look very much like the results of conformity.
We therefore seek out subsets of the products on which 
the two effects might be distinguishable, and the argument
above suggests starting with products that exhibit high levels
of individual variation in \smarks.

In particular, we associate with each product the 
{\em variance} of the \smarks assigned to it by
all its reviews.
We then group \products by variance, and perform the
signed-difference analysis above on sets of \products having
fixed levels of variance.
We find (Figure~\ref{fig:signed-diff-var}) that the effect
of signed difference to the average changes smoothly
but in a complex fashion as the variance increases.
The role of variance can be summarized as follows.
\begin{itemize}
\item When the variance is very low, the reviews with
the highest helpfulness ratios are those with the average \smark.
\item With moderate values of the variance, the reviews
evaluated as most helpful are those that are slightly above
the average \smark.
\item As the variance becomes large, reviews with \smarks both above
{\em and} below the average are evaluated as more helpful
than those that have the average \smark (with the positive reviews
still deemed somewhat more helpful).
\end{itemize}
These principles suggest some qualitative ``rules'' for how ---
all other things being equal --- one can seek good 
helpfulness
evaluations in our setting: With low variance go with the average;
with moderate variance be slightly above average;
and with high variance avoid the average.

This qualitative enumeration of principles
initially seems to be fairly elaborate;
but as we show in Section~\ref{sec:gaussians}, all these principles are 
consistent with a simple model of individual bias in
the presence of controversy.
Specifically, suppose that opinions are drawn from a mixture
of two \unimodalinformal distributions --- one with larger mixing weight
whose mean is above the overall mean of the mixture, 
and one with smaller mixing weight whose mean is below it.
Now suppose that each user has an opinion from this mixture, corresponding
to their own personal 
score for the product, and they evaluate
reviews as helpful if the review's \smark is within some
fixed tolerance of their own.
We can show that in this model, as variance increases from $0$,
the reviews evaluated as most helpful are initially slightly
above the overall mean, and eventually a ``dip'' in \helpfulness
appears around the mean.

Thus, a simple model can in principle account for 
the fairly complex series of effects illustrated in 
Figure~\ref{fig:signed-diff-var}, and provide a 
hypothesis for an underlying mechanism.
Moreover, the effects we see are surprisingly robust 
as we look at different national Amazon sites for
the U.K., Germany, and Japan.  
Each of these communities has evolved independently, 
but each exhibits the same set of patterns.
The one non-trivial and systematic deviation from the pattern
among these four countries is in the analogue of
Figure~\ref{fig:signed-diff-var} for Japan: as with the
other countries, a ``dip'' appears at the average in
the high-variance case, but in Japan the portion of the
curve {\em below} the average is higher.
This would be consistent with a version of our
two-distribution individual-bias model in which the
distribution below the average has higher mixing weight ---
representing an aspect of the brilliant-but-cruel hypothesis
in this individual-bias framework, and only for this one
national version of the site.

\xhdr{Controlling for text: Taking advantage of \plagiarism}
Finally, we return to one further issue discussed earlier:
how can we offer evidence that these non-textual features
aren't simply serving as correlates of review-quality
features that are intrinsic to the text itself?
In other words, are there experiments that can address
the \contenthypprefix straw man hypothesis above?

To deal with this, we make use of rampant \plagiarism and
duplication of reviews on Amazon.com (the causes and
implications of this phenomenon are beyond the scope of this
paper).  This is a fact that has been noted and studied by
earlier researchers \cite{David+Pinch:06a}, and for most
applications it is viewed as a pathology to be remedied.
But for our purposes, it makes possible a remarkably
effective way to control for the effect of review text.
Specifically, we define a {\em \plagiarized pair} of reviews
to be two reviews of different products with near-complete
textual overlap, and we enumerate the several thousand
instances of plagiarized pairs on Amazon.  (We distinguish
these from reviews that have been cross-posted by Amazon
itself to different versions of the same product.)

Not only are the two members of a \plagiarized pair
associated with different products; very often they also
have significantly different \smarks and are being used on
products with different averages and variances.  (For
example, one copy of the review may be used to praise a book
about the dangers of global warming while the other copy is
used to criticize a book that is favorable toward the oil
industry).  
We find significant differences in the \hratios within 
plagiarized pairs, and these differences confirm
many of the the effects we observe on the full dataset.
Specifically, within a \plagiarized pair, the copy of the
review that is closer to the average gets the higher \hratio
in aggregate.

Thus the widespread copying of reviews provides us with
a way to see that a number of social feedback effects ---
based on the 
score of a review and its relation to other scores ---
lead to different outcomes even for reviews that are 
textually close to identical.

\xhdr{Further related work}
We also mention some relevant prior literature that
has not already been discussed above. 
The role of social and cognitive factors in 
purchasing decision-making has
been extensively studied in psychology and marketing
\citep{Cialdini+Goldstein:04a,Escalas+Bettman:05a,Fitzsimons+al:02a,Underhill:99a},
recently making use of brain imaging methodology \citep{Knutson+al:07}.
Characteristics of the distribution 
of review \smarks (which differ from \hvotes) on Amazon and
related sites have been studied previously
\citep{Chevalier+Mayzlin:06a,Hu+Pavlou+Zhang:06a,Wu+Huberman:08c}.
Categorizing text by quality has 
been proposed for a number of 
applications \citep{Jindal+Liu:07a,Sen+al:07a,Agichtein+al:08a,Harper+al:08a}.
Additionally, 
our notion of variance is potentially related to 
the idea that people play different roles 
in on-line discussion  \citep{Welser+al:07a}.

\section{Data}
\label{sec:data}

Our experiments employed a dataset of over 4 million
Amazon.com book reviews (corresponding to 
roughly 675,000
books), 
of which more than 1 million received at least 10
\hvotes each.
We made extensive use of the Amazon Associates Webservice
(AWS) API to collect this data.\footnote{We used the AWS API
  version 2008-04-07.  Documentation is available at
  \url{http://docs.amazonwebservices.com/AWSECommerceService/2008-04-07/DG/}
  .}
We describe the process in this section, with particular
attention to measures we took to avoid sample bias.

We would ideally have liked to work with {\em all} book
reviews posted to Amazon. However, one can only access
reviews via queries specifying particular books
by their
Amazon product ID,
 or ASIN (which is the same as ISBN for most
books),
and
we are not aware of any publicly available list of
all Amazon book ASINs.
However, the API allows one to query for books in a specific category (called
a {\em browse-node} in AWS parlance and corresponding to a section on the
Amazon.com website),
and the best-selling titles up to a limit of 4000 in each browse-node can be obtained in this way.

To create our initial list of books, therefore, we performed
queries for all 3855 categories three levels deep in the
Amazon browse-node hierarchy (actually a directed acyclic graph) rooted at
``Books$\rightarrow\,$Subjects''.  
An example
category is
  {\newcommand{\abnsep}{$\rightarrow\,$}
    \newcommand{\abn}[3]{\emph{#1\abnsep#2\abnsep#3}\xspace}
    \abn{Children's Books}{Animals}{Lions, Tigers \& Leopards}.
  } 
These queries resulted in the initial set of 3,301,940 
books,
where we count books listed in multiple categories only once.

We then performed a book-filtering step to deal with
``cross-posting'' of reviews across versions.
When Amazon
carries different versions of the same item --- for example,
different editions of the same book, including hardcover and
softcover editions and audio-books --- the reviews written
for
all versions are merged and displayed together on each
version's product page
and 
likewise 
returned by the API upon queries for any individual version.%
\footnote{At the time of data collection, the API did
  not provide an option to trace a review to a particular
  edition for which it was originally posted, despite the
  fact that the Web store front-end has included such links for
  quite some time.}
This means 
that multiple copies of the same
review exist for ``mechanical'', as opposed to user-driven,
reasons.\footnote{We make use of human-instigated review
  copying later in this study.}
To avoid including mechanically-duplicated reviews, we
retained only one of the set of alternate versions for each
book (the one with the most complete metadata).

The above process gave us a list of 674,018 books for which
we 
retrieved
reviews by querying AWS.  Although AWS
restricts the number of reviews returned for any given
product query to a maximum of 100, it 
turned out that 99.3\%
of our books had 100 or fewer reviews.
In the case of the remaining 4664 books,
we chose to retrieve the 100
{\em earliest} reviews for each product to be able to
reconstruct the information available to the authors and
readers of those reviews
to the extent possible. 
(Using the earliest reviews ensures
the reproducibility of our results, since the 100 earliest
reviews comprise a static set, unlike the 100 most helpful
or recent reviews.)
As a result, we ended up with 4,043,103 reviews;
although some
reviews were not retrieved due to the 
100-reviews-per-book API cap, 
the number of missing reviews averages out to roughly just one per ASIN
queried.
Finally, we focused on the 1,008,466 reviews that had at least 10 \hvotes each.

The size of our dataset compares favorably to that of
collections used in other studies looking at \hvotes:
\citet{Liu+al:07a} used about 23,000 digital camera reviews
(of which a subset of around 4900 were subsequently given
new \hvotes and studied more carefully);
\citet{Zhang+Varadarajan:06a} used about 2500 reviews of
electronics, engineering books, and PG-13 movies after
filtering out duplicate reviews and reviews with no more
than 10 \hvotes;
\citet{Kim+al:06a} used about 26,000 MP3 and digital-camera reviews after filtering of
duplicate versions and duplicate reviews and reviews with fewer than
5 \hvotes; 
and
\citet{Ghose+Ipeirotis:08a} considered ``all reviews since the product
was released into the market'' (no specfic number is given)
for about 400 popular audio and video players, digital cameras, and
DVDs.

\section{Effects of \headerformat{\deviation} from average and
  \headerformat{\variance}}
 \label{sec:variance}

Several of the \hypotheses that we have described concern
the relative position of an opinion about an entity
vis-\`a-vis the average opinion about that entity.  We now
turn, therefore, to the question of how the \hratio of a review
depends on its \smark's \deviation from the average \smark for
all reviews of the same book.  According to the
\conformityhyp,
the
 \hratio should be {\em lower} for reviews
with \smarks \emph{ either above or below} the \product
average, whereas the \brillianthyp translates to the
``asymmetric'' prediction that 
the
\hratio should be {\em
  higher} for reviews with \smarks \emph{below} the \product
average than for overly positive reviews.  (No specific
predictions for \hratio vis-\`a-vis \product average is
made by  either the \biasprefix or \contenthyp without further
assumptions about the distribution of individual opinions or
text quality.)

\xhdr{Defining the average}
For a given review, let the {\em \avgmarkcomp}
(abbreviation: {\em \avgmarkcompabbrev})
be the average \smark as computed over all reviews of that
\product in our dataset.

{This differs in principle from the {\em \avgmarkdisplayed}
  (abbreviation: {\em \avgmarkdisplayedabbrev}), the
  ``Average Customer Review'' score
  that Amazon itself displayed for the book at the time we
  downloaded the data.  One reason for the difference is
  that Amazon rounds the \avgmarkdisplayedabbrev to the
  nearest half-star (e.g., 3.5 or 4.0) --- but for our
  experiments it is preferable to have a greater degree of
  resolution.  Another possible source of difference is the
  very small 
(0.7\%)
fraction of books,
  mentioned in Section \ref{sec:data}, for which the entire
  set of reviews could not be obtained via AWS: the
  \avgmarkdisplayedabbrev would be partially based on
  reviews that came later than the first 100 and which would
  thus not be in our dataset. 
  However, the mean absolute difference between the
  \avgmarkcompabbrev when rounded to the nearest half-star
  (0.5 increment) and the \avgmarkdisplayedabbrev is only
  0.02.

Note that both scores can differ from the ``Average Customer Review''
score that Amazon displayed {\em at the time a \hevaluator provided
their \hvote}, since this time might pre-date some of the reviews
for the book that are in our dataset (and hence that Amazon based its
\avgmarkdisplayedabbrev on). In the absence of timestamps on \hvotes,
this is not a factor that can be controlled 
for.

\xhdr{\Deviation experiments}

\begin{figure}[t]
\begin{center}
\includegraphics[width=3in,viewport= 50 200 550 600,clip]{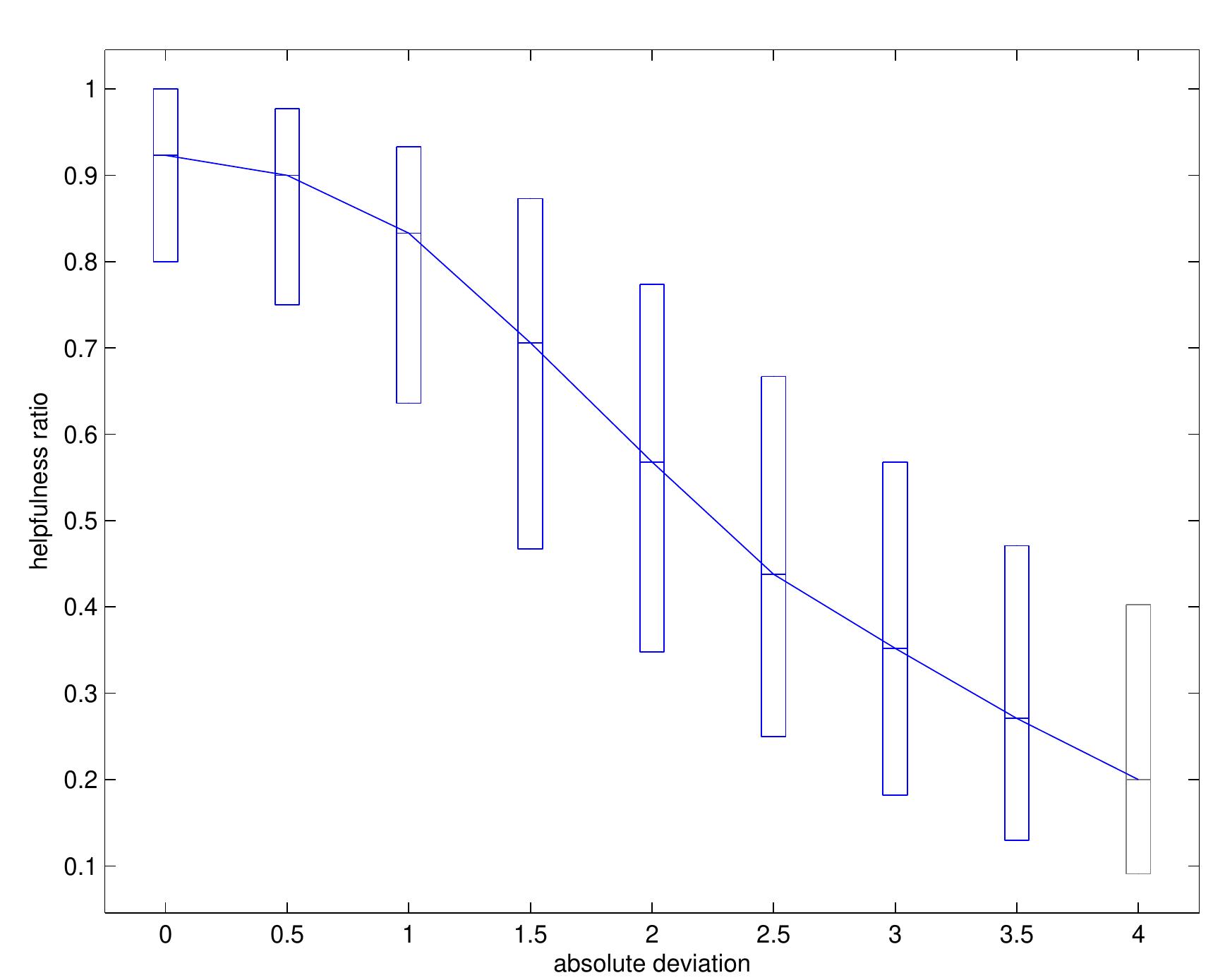}
\caption{
\label{fig:unsigned-diff-novar} \Hratio declines with
 the absolute value of a review's \deviation 
 from the \avgmarkcompabbrev; this behavior is predicted by
 the \conformityhyp but not ruled 
out by the other
 hypotheses. 
\newline
\hspace*{.1in} The line segments within the
 bars (connected by the descending line) indicate the median
 \hratio; the 
bars depict
 the \hratio's second and third quantiles. 
\newline
\hspace*{.1in}Throughout, grey bars indicate that the amount of data at that $x$ value represents $.1\%$ or less of the data depicted in the plot. 
}
\end{center}
\end{figure}

We first check the prediction of the \conformityhyp that the
\hratio of a review will vary inversely with the absolute
value of the difference between the review's \smark and the
\avgmarkcomp --- we call this difference the review's {\em
  \deviation}.

Figure \ref{fig:unsigned-diff-novar} indeed shows a very strong
inverse correlation between the median \hratio 
and the absolute \deviation
, as predicted by the \conformityhyp.  However, this data
does not completely disprove the \brillianthyp, since for a
given absolute \deviation $|x|>0$, it could conceivably happen
that reviews with positive \deviations $|x|$ (i.e. more
favorable than average) could have much worse \hratios than
reviews with negative \deviation $-|x|$, thus dragging down the
median \hratio.
Rather, to directly assess
the \brillianthyp, 
we must consider {\em
\sdeviation}, not just absolute \deviation.

\begin{figure}[ht]
\begin{center}
  \includegraphics[width=3in, viewport= 50 200 550 600,clip]{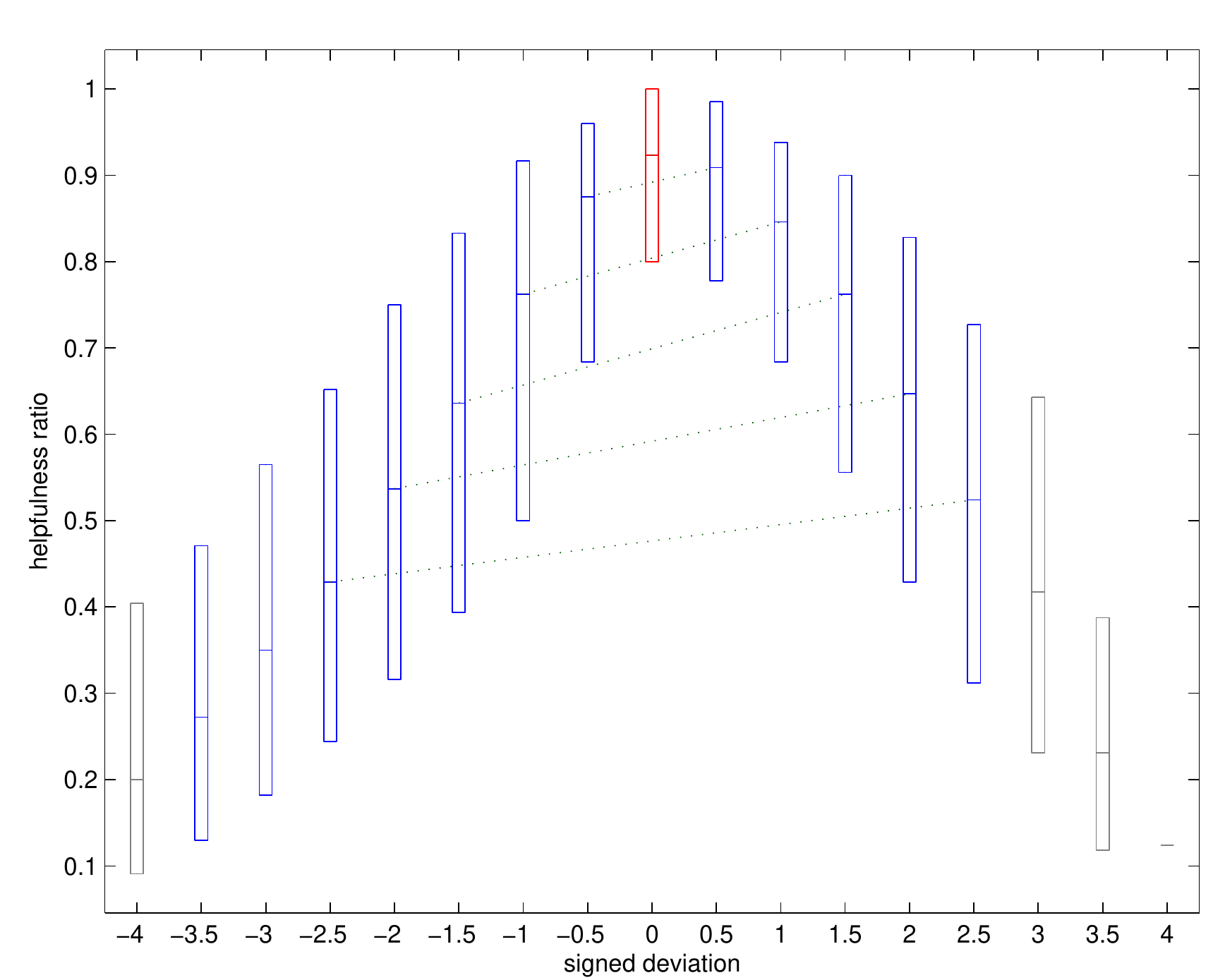}
  \caption{\label{fig:signed-diff-novar} The dependence of
    \hratio on a review's \sdeviation from average is
    inconsistent with both the \brilliantprefix and, because of
    the asymmetry, the \conformityhyp.  
  }
\end{center}
\end{figure}

Surprisingly, the effect of \sdeviation on median \hratio,
depicted in the ``Christmas-tree'' plot of Figure
\ref{fig:signed-diff-novar}, turns out to be different from
what {\em either} hypothesis would predict.

The \brillianthyp clearly does not hold for our data:
among reviews with the same absolute \deviation $|x|>0$, the
relatively positive ones (\sdeviation $|x|$) generally have a higher
median \hratio than the relatively negative ones (\sdeviation of $-|x|$), as
depicted by the positive slope of the
{green dotted} lines connecting ($-|x|$,$|x|$)
pairs of datapoints.

But Figure \ref{fig:signed-diff-novar} {\em also} presents
counter-evidence for the \conformityhyp, since
that hypothesis incorrectly predicts that the connecting
lines would be horizontal.

\begin{figure*}[t]
  \begin{center}
    \includegraphics[height=4in, viewport= 50 200 550 600,clip]{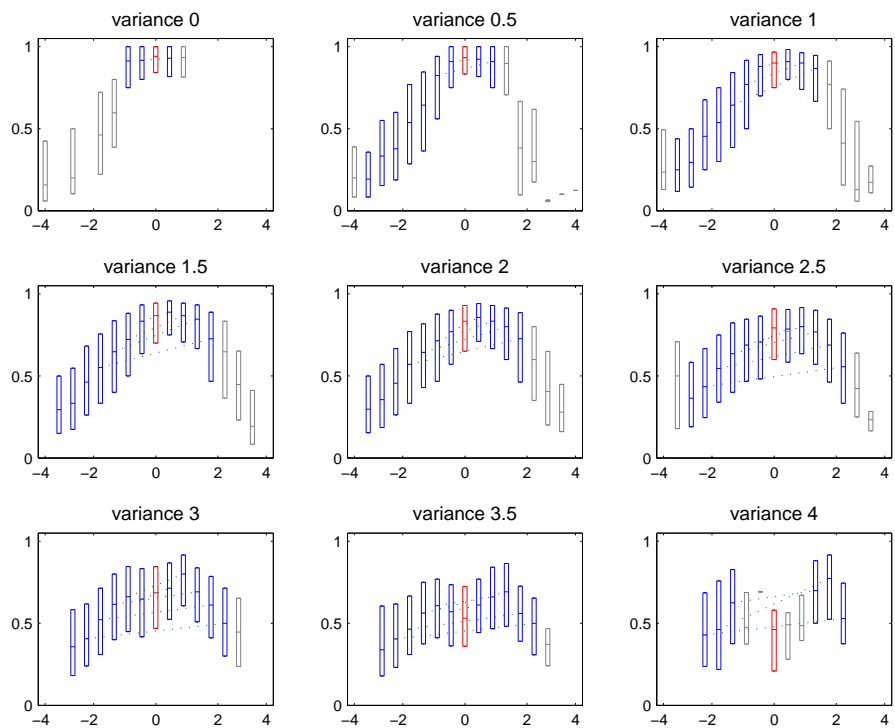}
    \caption{\label{fig:signed-diff-var} As the variance of the \smarks of reviews for a particular product increases, the median \hratio curve becomes two-humped and the \hratio at signed \deviation 0 (indicated in red) no longer represents the unique global maximum. There are non-zero \sdeviations in the plot for \variance 0 because we rounded \variance values to the nearest .5 increment.}
\end{center}
\end{figure*}

To account for Figure \ref{fig:signed-diff-novar}, one could simply impose
upon the \conformityhyp an extra ``tendency towards positivity''
factor, but this would be quite unsatisfactory: it wouldn't
suggest any underlying mechanism for this factor. So, we turn to the
\biashyp instead.

In order to distinguish between \conformity
effects and  \biasprefix effects, we need to examine cases in which 
individual people's opinions do {\em not} come from exactly the same
(\unimodalinformal, say) 
distribution; for otherwise, the composite of their
\biases could produce \hratios
that look very much like the results of \conformity.
One natural place to begin to seek settings in which \bias and \conformity are
distinguishable, in the sense just described, is in cases in which
there is at least high {\em \variance} in the \smarks.
Accordingly, Figure \ref{fig:signed-diff-var} separates \products by
the variance of the \smarks in the reviews for that \product  in our dataset.

One can immediately observe some striking effects of \variance.  First,
we see that as variance increases, the ``camel plots''  of Figure
\ref{fig:signed-diff-var} go from 
a single hump to two.\footnote{This is a reversal of nature, where 
Bactrian (two-humped) camels are more agreeable than one-humped
Dromedaries.}
We also note that while in the previous figures it was the reviews
with a \sdeviation of exactly zero that had the highest \hratios,
here we see that once the \variance among reviews for a \product is
3.0 or greater, the highest \hratios are clearly achieved for \products with
\sdeviations close to but still noticeably above
zero. (The
beneficial effects of having a \smark slightly above the mean are
already discernible, if small, at variance 1.0 or so.)

Clearly, these results indicate that \variance is a key factor that
any hypothesis needs to incorporate.
In Section \ref{sec:gaussians}, we develop a simple \biasprefix model 
that does so; but first, there is one last hypothesis that we need to consider.

\vspace*{.6in}
\section{Controlling for text quality: experiments with
  \headerformat{\plagiarism}}

As we have noted, our analyses do not explicitly take into
account the actual text of reviews.  It is not impossible,
therefore, that review text quality may be a confounding
factor and that our straw-man \contenthyp might hold.  
Specifically, we have shown that \hratios 
appear to be dependent on two key non-textual 
aspects of reviews, namely, 
on
\deviation from the \avgmarkcompabbrev and
 on \smark
\variance within reviews for a given product; but we have not shown
that our results are not simply explained by review quality.

Initially, it might seem that the only way to control for text quality
is to read a sample of reviews and determine whether the Amazon
\hratios assigned to these reviews are accurate.  Unfortunately, it
would require a great deal of time and human effort to gather a
sufficiently large set of re-evaluated reviews, and human
re-evaluations can be subjective; so it would be preferable to find a
more efficient and objective procedure.
\footnote{
\citet{Liu+al:07a} did perform a manual re-evaluation of 4909 digital-camera reviews, finding that the original \hratios did not seem well-correlated with the stand-alone
comprehensiveness of the reviews.  But note that this could just mean that at least some of the original \hevaluators were using a different
standard of text quality 
(Amazon does not specify any particular standard or definition of \helpfulness).
Indeed, the exemplary ``fair''
review quoted by Liu et al. begins, ``There is nothing wrong with
the [product] except for the very noticeable delay between
pics. [Description of the delay.] Otherwise, [other aspects] are fine
for anything from Internet apps to ... print enlarging.  It is
competent, not spectacular, but it gets the job done at an agreeable
price point.''  Liu et al. give this a rating of ``fair'' because it only
comments on some of the product's aspects, but the Amazon \hevaluators
gave it a \hratio of 5/6, which seems reasonable.
Also, reviews might also be evaluated vis-\`a-vis the totality of all reviews, 
i.e.,  a review might be rated helpful if it provides complementary
information or ``adds value''.
For instance, 
a one-line review that points out a serious flaw in
another review could well be considered ``helpful'',  but would not
rate highly under Liu et al.'s scheme.  

It is also worth pointing out subjectiveness can remain an issue even
with respect to a given text-only evaluation scheme.
The two human re-evaluators who used Liu et al.'s
\citeyearpar{Liu+al:07a} standard
assigned different \helpfulness categories (in a four-category
framework) to  619=12.5\% of the reviews considered, indicating that
there can be substantial subjectiveness involved in determining review
quality even when a single standard is initially agreed upon.}

A different potential approach would be to use machine learning to train
an algorithm to 
automatically determine the degree of \helpfulness of each review.
Such an
approach would indeed involve less human effort, and could thus be
applied to larger numbers of reviews.  
However, we could not draw the
conclusions we would want to: any mismatch between the predictions of
a trained classifier and the \hratios observed in held-out
reviews could be attributable to errors by the algorithm, rather than
to the actions of the Amazon \hevaluators.\footnote{
\citet{Ghose+Ipeirotis:08a} observe that their trained classifier
often performed poorly for reviews of products with ``widely
fluctuating'' \smarks, and explain this with an assertion that
the Amazon \hevaluators are not judging text quality in such situations. But
there is no evidence provided to dismiss the alternative hypothesis that the
\hevaluators are correct and that, rather, the algorithm makes mistakes because
reviews are more complex in such situations and the classifier uses
relatively shallow textual features. 
 
} 

We thus find ourselves in something of a quandary: we seem to lack any
way to derive a sufficiently large set of objective and accurate
re-evaluations of  \helpfulness.
Fortunately, we can bring to bear on this problem two key insights:
\begin{enumerate}
\item 
Rather than try to re-evaluate all reviews for their
\helpfulness, we can focus on reviews that are {\em guaranteed} to
have very similar levels of textual quality.  
\item Amazon data contains many instances of
nearly-identical reviews
\citep{David+Pinch:06a} --- 
and identical reviews must {\em
necessarily} exhibit the same level of text quality.
\end{enumerate}

Thus, in the remainder of this section, we consider whether the
effects we have analyzed above hold on pairs of {\em \plagiarized}
reviews.  

\xhdr{Identifying \plagiarism
 (as distinct from ``justifiable
copying'')} Our choice of the term \plagiarism is meant to be somewhat evocative, 
because we disregard several types of  arguably justifiable 
copying or duplication in which there is no overt attempt
to make the copied review seem to be a genuinely new piece of text;
the reason is because this kind of copying does not suit our purposes.
However, ill intent cannot and should not be ascribed to the authors of the remaining reviews; we have attempted to indicate this by the inclusion of scare quotes
around the term.

In brief, we only considered pairs of
reviews where the two reviews were posted to different books --- this avoids various types of relatively obvious self-copying (e.g., where an author reposts a review under their user ID after initially posting it anonymously), since obvious copies might be evaluated differently.

We next adapted the code of 
\citet{Sorokina+al:06a} to identify 
those pairs of reviews of different \products that have highly similar text.
To do so, we needed to decide on 
a similarity threshold that determines whether or not we deem a review
pair to be \plagiarized.
A reasonable option would have been  to consider only reviews with
identical text, which would ensure that the reviews in the pairs
had exactly the same \textualquality.  However,  since the reviews in
the analyzed pairs are posted for different products,  it is normal to
expect that some authors modified or added to the text of the original
review 
to make the \plagiarized copy better fit its
new context.
For this reason, we employed a 
threshold of $70\%$ or more nearly-duplicate sentences,
where near-duplication was measured via the code of
\citet{Sorokina+al:06a}.\footnote{\citet{Kim+al:06a}, who also noticed that
the phenomenon of review alteration affected their attempts to remove
duplicate reviews, used a similar threshold of 80\% repeated bigrams.}
This yielded $8,313$ \plagiarized pairs;
an example is shown in Figure \ref{fig:plagiarism_example}.
Manual inspection of a sample revealed that the review pairs captured by our
threshold indeed seem to consist of close copies.

\begin{figure}[h]
\begin{center}
\begin{minipage}{3.35in}{\includegraphics[width=3.35in]{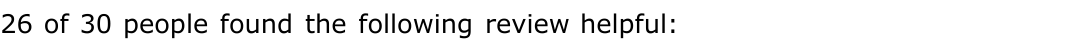}}
\end{minipage}\\
\begin{minipage}{3.35in}{\includegraphics[width=3.35in]{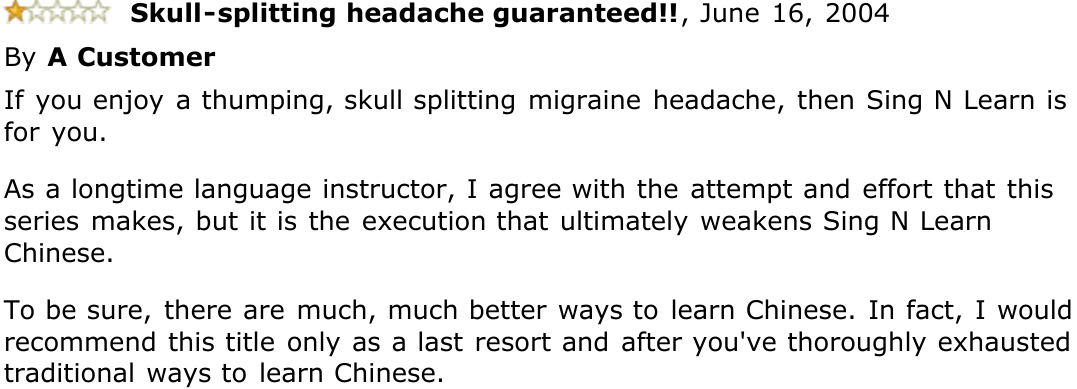}}\hspace*{-1.9in}$\cdots$
\end{minipage}\\
\begin{minipage}{3.35in}{\includegraphics[width=3.35in]{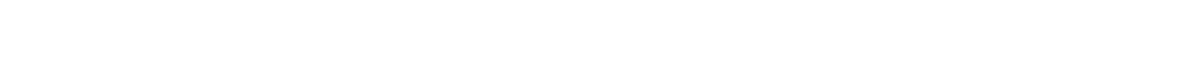}}
\end{minipage}\\
\begin{minipage}{3.35in}{\includegraphics[width=3.35in]{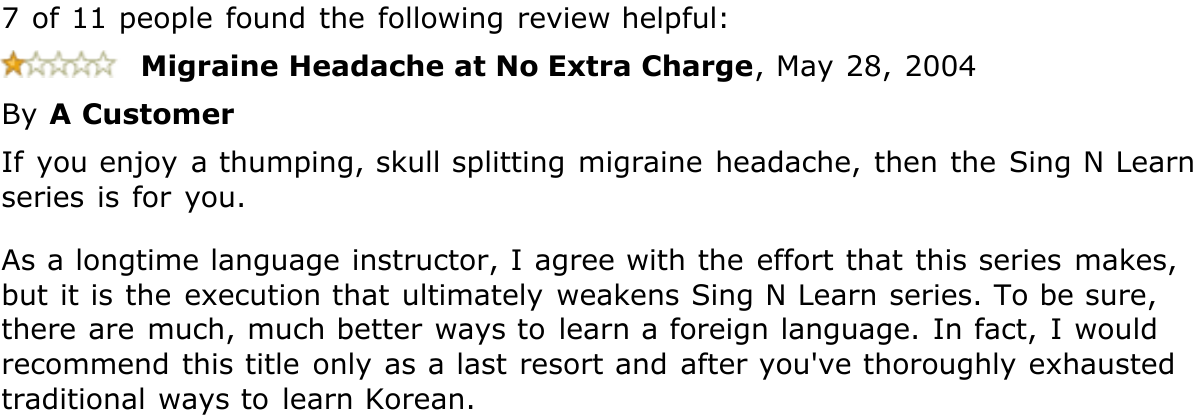}}\hspace*{-2in}$\cdots$
\end{minipage}
\caption{The first paragraphs of \plagiarized reviews posted for the \products
\texttt{Sing 'n Learn Chinese} and \texttt{Sing 'n Learn Korean}. In the second review, the title is different and
the word ``chinese'' has been replaced by ``korean'' throughout.  
Sources: {\footnotesize \protect\url{http://www.amazon.com/review/RHE2G1M8V0H9N/ref=cm_cr_rdp_perm}}
and 
\protect\url{http://www.amazon.com/review/RQYHTSDUNM732/ref=cm_cr_rdp_perm}.
\label{fig:plagiarism_example}}
\end{center}
\end{figure}

\vspace*{0.3in}

\xhdr{Confirmation that \textualquality is not the (only) explanatory factor} 
Since for a given pair of \plagiarized reviews the \textualquality of
the two copies should be essentially the same, a statistically
significant difference between the \hratios of the members of such
pairs is a strong indicator of the influence of a non-textual factor
on the \hevaluators.  

An initial test of the data reveals that the mean difference in
\hratio between \plagiarized copies is 
very close to zero.
However, a confounding factor is that for many of our
pairs, the two copies may occur in contexts that are practically
indistinguishable.  Therefore, we bin the pairs by how different their
absolute \deviations are, and consider whether \hratios differ at
least for pairs with very different
\deviations. 
More formally, for $i,j \in \{0,0.5,\cdots,3.5\}$ where $i < j$, 
we write $i$\moreh$j$ (conversely, $i$\lessh$j$) when the \hratio of reviews
with absolute \deviation $i$ is \significantly larger (conversely, smaller)
than that for reviews with absolute \deviation $j$. 
Here, {\em \significantly
larger} or {\em smaller} means that the Mantel-Haenszel test
for whether the \helpfulness odds ratio is equal to 1 returns 
a $95\%$ confidence interval
that does not contain 1.\footnote{To avoid drawing conclusions based on possible
numerical-precision inaccuracies, we consider any confidence interval that overlaps the interval
[0.995,1.005] to contain 1.  This ``overlap'' policy affects only two bins in Table
\ref{tab:plagiarism_unsigned} and two bins in Table
\ref{tab:plagiarism_signed}.}
The Mantel-Haenszel test \cite{Agresti:96a} measures the strength of
association between two groups, giving more weight to groups with more data.
(Experiments with an alternate empirical sampling test were consistent.)
We disallow $j= 4$ since
there are only relevant 24 pairs which would have to be distributed among
$8$ $(i,j)$ 
bins.

The existence of even a {\em single} pair $(i,j)$ in which $i$\moreh$j$ or  $i$\lessh$j$
would already be inconsistent with the \contenthyp. Table
\ref{tab:plagiarism_unsigned} shows  that in fact, there is a
\significant difference in a {\em large
majority} of cases.
Moreover, we see no ``\lessh'' symbols; this  is consistent
with
Figure \ref{fig:unsigned-diff-novar}, which showed that the \hratio is
inversely correlated with absolute \deviation prior to controlling for
\textualquality.

We also binned the pairs by {\em signed} \deviation.
The results, shown in Table
\ref{tab:plagiarism_signed}, are consistent with Figure \ref{fig:signed-diff-novar}.
First, all but one of the statistically significant results indicate that
\plagiarized reviews with \smark closer to the \product average are
judged to be more helpful.
Second, the ($-i,i$) results are consistent with the asymmetry
depicted in Figure \ref{fig:signed-diff-novar} (i.e., the ``upward slant'' of
the green lines).

Note that the sparsity of the 
\plagiarism data precludes an 
analogous investigation
of \variance as 
as a
contextual factor.

\newcommand{\ci}{\textcolor{white}} 

\begin{table}[htdp]
\begin{center}
{\small
\begin{tabular}{|c||c|c|c|c|c|c|c|}\hline \backslashbox{$i$}{$j$} & 0.5 & 1 &  1.5 & 2 & 2.5 & 3 & 3.5 \\\hline \hline
0 & \ci\lessh & \moreh &\moreh & \moreh & \moreh & \moreh & \moreh \\
\hline 0.5 &  \ec & \moreh & \moreh & \morehu & \moreh & \moreh & \moreh \\
\hline 1 & \ec &\ec  & \moreh & \morehu & \moreh & \moreh & \moreh \\
\hline 1.5 & \ec &\ec  &\ec  & \lesshu & \moreh & \moreh & \moreh \\
\hline 2 & \ec & \ec &\ec  & \ec & \lesshu & \moreh & \moreh \\
\hline 2.5 & \ec & \ec & \ec &\ec  & \ec & \moreh & \moreh \\
\hline 3 &\ec  & \ec &\ec  & \ec & \ec &\ec  & \ci\moreh \\\hline
\end{tabular}
}
    \caption{\label{tab:plagiarism_unsigned} \Plagiarized reviews with a lower absolute \deviation
    tend to have larger \hratios than duplicates with higher
    absolute deviations.  Depicted:
whether reviews with
    deviation $i$ have an \hratio significantly larger (\moreh) or significantly
    smaller (\lessh, no such cases) than duplicates with absolute
    \deviation $j$ (blank: no significant difference).
}
\end{center}
\end{table}

\newtheorem{theorem}{Theorem}[section]
\newcommand{\xproof}[1]{
{\noindent {\it Proof.} {#1} \rule{2mm}{2mm} \vskip \belowdisplayskip} }
\def\bal{{p}}
\def\cbal{{q}}
\def\controv{{\alpha}}
\def\eps{{\varepsilon}}

\section{A Model Based on \headerformat{\bias} and Mixtures of Distributions}
\label{sec:gaussians}

We now consider how the main findings about
helpfulness, variance, and divergence from the mean 
are consistent with a simple model based on individual
bias with a mixture of opinion distributions.
In particular, our model exhibits the 
phenomenon observed in our data that increasing
the variance shifts the helpfulness distribution
so it is first unimodal 
and subsequently (with larger variance) develops 
a local minimum around the mean.

The model assumes that
helpfulness evaluators can come from two different
distributions: one consisting of evaluators who are
positively disposed toward the product, and 
the other consisting of evaluators who are negatively
disposed toward the product.
We will refer to these two groups as the {\em positive} and 
{\em negative} evaluators respectively.

We need not make specific distributional assumptions about
the evaluators; rather, we simply assume that their opinions
are drawn from {\em some} underlying distribution with
a few basic properties.
Specifically, let us say that a function $f : {\bf R} \rightarrow {\bf R}$
is {\em $\mu$-centered}, for some real number $\mu$,
if it is {\em unimodal} at $\mu$, centrally symmetric, 
and $C^2$ (i.e. it possesses a continuous second derivative).
That is, $f$ has a unique local maximum at $\mu$,
$f'$ is non-zero everywhere other than $\mu$,
and $f(\mu + x) = f(\mu - x)$ for all $x$.
We will assume that both positive and negative evaluators
have one-dimensional opinions drawn from 
(possibly different) distributions with density
functions that are $\mu$-centered for distinct values of $\mu$.

Our model will involve two parameters:
the {\em balance} between positive and negative reviewers $\bal$,
and a {\em controversy level} $\controv > 0$.
Concretely, we assume that there is a $\bal$ fraction of positive evaluators
and a $1 - \bal$ fraction of negative evaluators.
(For notational simplicity, we sometimes write $\cbal$ for $1 - \bal$.)
The controversy level controls the distance between the means
of the positive and negative populations:
we assume that for some number $\mu$, the density function $f$ for
positive evaluators is $(\mu + \cbal \controv)$-centered,
and the density function $g$ for negative evaluators is
$(\mu - \bal \controv)$-centered.
Thus, the density function for the full population is
$h(x) = \bal f(x) + \cbal g(x)$, and it has mean 
$\bal (\mu + \cbal \controv) + \cbal (\mu - \bal \controv) = \mu$.
In this way, our parametrization allows us to keep the
mean and balance fixed while observing the effects as we vary
the controversy level $\controv$.

Now, under our individual-bias assumption, we posit that 
each helpfulness evaluator has an opinion $x$ drawn from $h$, 
and each regards a review as helpful if it expresses an
opinion that is within a small tolerance of $x$.
For small tolerances, we expect therefore that
the helpfulness ratio of reviews giving a score of $x$, 
as a function of $x$, can be approximated by $h(x)$.
Hence, we consider the shape of $h(x)$ and ask whether
it resembles the behavior of helpfulness ratios observed in the real data.

Since the controversy level $\controv$ in our model affects the
variance in the empirical data ($\controv$ is the distance between the
peaks of the two distributions, and is thus related to the variance,
but the balance $\bal$ is also a factor),
we can hope that at as $\controv$ 
increases one obtains qualitative properties consistent with the data:
first a unimodal distribution with peak between the means of $f$ and $g$, and
then a local minimum near the mean of $h$.
In fact, this is precisely what happens.
The main result is the following.

\newcommand{\ms}{\textcolor{white}{-}}

\begin{table}[tdp]
\begin{center}
{\small
\begin{tabular}{|c||c|c|c|c|c|c|c|}\hline 
\backslashbox{$i$}{$j$} & -0.5 & -1 &  -1.5 & -2 & -2.5 & -3 & -3.5 \\\hline \hline
0 &                  \ci\lessh & \moreh & \moreh & \moreh & \moreh & \moreh & \moreh \\
\hline -0.5   &  \ec        & \moreh & \moreh & \lesshu & \moreh & \moreh & \lesshu \\
\hline -1 &       \ec        &\ec         & \moreh & \lesshu & \lesshu & \moreh & \lesshu \\
\hline -1.5 & \ec &\ec  &\ec                          & \lesshu & \moreh & \lesshu & \lesshu \\
\hline -2 & \ec & \ec &\ec  & \ec                                    & \lesshu & \lesshu & \lesshu \\
\hline -2.5 & \ec & \ec & \ec &\ec  & \ec                                        & \moreh & \moreh \\
\hline -3 &\ec  & \ec &\ec  & \ec & \ec &\ec                                                    & \ci\moreh \\\hline
\multicolumn{7}{c}{} \\
\hline \backslashbox{$i$}{$j$} & \ms0.5 & \ms1 &  \ms1.54 & \ms2 & \ms2.5 & \ms3 & \ms3.5 \\\hline \hline
0 &                          \lesshu & \lesshu & \moreh & \moreh & \lesshu & \lesshu & \lesshu \\
\hline 0.5 &  \ec                  & \moreh & \moreh & \lesshu & \lesshu & \lesshu & \lesshu \\
\hline 1 & \ec &\ec                              & \moreh & \lessh & \lesshu & \lesshu & \lesshu \\
\hline 1.5 & \ec &\ec  &\ec                                 & \lesshu & \lesshu & \lesshu & \lesshu \\
\hline 2 & \ec & \ec &\ec  & \ec                                          & \lesshu & \lesshu & \lesshu \\
\hline 2.5 & \ec & \ec & \ec &\ec  & \ec                                             & \lesshu & \lesshu \\
\hline 3 &\ec  & \ec &\ec  & \ec & \ec &\ec                                                        & \lesshu \\\hline
\multicolumn{7}{c}{} \\
\hline \backslashbox{}{$i$} & -0.5 & -1 &  -1.5 & -2 & -2.5 & -3 & -3.5 \\\hline \hline
$-i$                        &        & \moreh & \moreh &   &        &      &    \\\hline
\end{tabular}
}
\end{center}

\caption{\label{tab:plagiarism_signed}
The same type of analysis as Table \ref{tab:plagiarism_signed} but
with {\em signed} deviation.
The first (resp. second) table is consistent with the lefthand (resp. righthand) side of Figure \ref{fig:signed-diff-novar}.  The third table is consistent with the ``upward slant'' of the green lines in Figure \ref{fig:signed-diff-novar}: for the same absolute \deviation value, when there is a  significant difference in \helpfulness odds ratio, the difference is in favor of the positive \deviation. \newline
\hspace*{.2in}{\small (There are a noticeable number of blank cells, indicating that a statistically
significant difference was not observed for the corresponding bins, 
due
to sparse data issues: there are twice as many bins as in the absolute-deviation analysis but
the same number of pairs.)}
}
\end{table}

\begin{theorem}
For any choice of $f$, $g$, and $\bal$
as defined as above, 
there exist positive constants $\eps_0 < \eps_1$ such that
\begin{itemize}
\item[(i)] When $\controv < \eps_0$, the combined density 
$h(x)$ is unimodal, with maximum strictly between 
the mean of $f$ and the mean of $g$.
\item[(ii)] When $\controv > \eps_1$, the combined density function
$h(x)$ has a local minimum between the means of $f$ and $g$.
\end{itemize}
\label{thm:general-mixture}
\end{theorem}
\xproof{
We first prove (i).
Let us write $\mu_f = \mu + \cbal \controv$ for the mean of $f$, and
$\mu_g = \mu - \bal \controv$ for the mean of $g$.
Since $f$ and $g$ have unique local maxima at their means,
we have $f''(\mu_f) < 0$ and $g''(\mu_g) < 0$.
Since these second derivatives are continuous, there exists
a constant $\delta$ such that 
$f''(x) < 0$ for all $x$ with $|x - \mu_f| < \delta$, and
$g''(x) < 0$ for all $x$ with $|x - \mu_g| < \delta$.
Since $\mu_f - \mu_g = \alpha$, 
if we choose $\alpha < \delta$, then 
$f''(x)$ and $g''(x)$ are both strictly negative over the entire
interval $[\mu_g, \mu_f]$.

Now, $f'(x)$ and $g'(x)$ are both positive for $x < \mu_g$, and
they are both negative for $x > \mu_f$.
Hence $h(x) = \bal f(x) + \cbal g(x)$ has the properties that
(a) $h'(x) > 0$ for $x < \mu_g$; (b) $h'(x) < 0$ for $x > \mu_f$,
and (c) $h''(x) < 0$ for $x \in [\mu_g, \mu_f]$.
From (a) and (b) it follows that $h$ must achieve its maximum in
the interval $[\mu_g, \mu_f]$, and from (c) it follows
that there is a unique local maximum in this interval.
Hence setting $\eps_0 = \delta$ proves (i).

For (ii), since $f$ and $g$ must both, as density functions that are
both 
centered around their respective means, go to $0$ as $x$ increases
or decreases arbitrarily, we can choose a constant $c$
large enough that 
$f(\mu_f - x) + g(x + \mu_g) < \min(\bal f(\mu_f), \cbal g(\mu_g))$
for all $x > c$.
If we then choose $\controv > c/\min(\bal,\cbal)$, we have
$\mu_f - \mu > c$ and $\mu - \mu_g > c$, and so 
$h(\mu) = \bal f(\mu) + \cbal g(\mu) 
\leq f(\mu) + g(\mu)
< \min(\bal f(\mu_f), \cbal g(\mu_g))
\leq \min(h(\mu_f), h(\mu_g))$,
where the second inequality follows from the definition of $c$
and our choice of $\alpha$.
Hence, $h$ is lower at its mean $\mu$ than at either of
$\mu_f$ or $\mu_g$, and hence it must have a local
minimum in the interval $[\mu_g,\mu_f]$.
This proves (ii) with $\eps_1 = c / \min(\bal,\cbal)$.
}

\begin{figure}[h]
  \centering
 \includegraphics[width=0.45\textwidth,viewport= 70 200 550 600,clip]{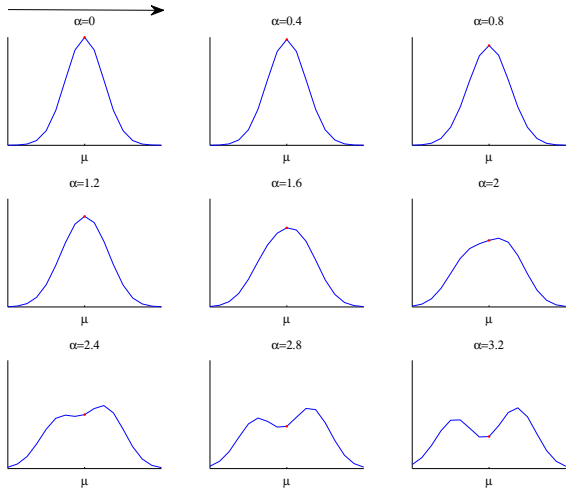} 
  \caption{An illustration of Theorem \ref{thm:symmetric-mixture}, using translated Gaussians as an example 
--- the theorem itself does not require any Gaussian assumptions. }
  \label{fig:gaussians-synthetic}
\end{figure}

For density functions at this level of generality,
there is not much one can say about the unimodal shape
of $h$ in part (i) of Theorem~\ref{thm:general-mixture}.
However, if $f$ and $g$ are translates of the same function,
and their next non-zero derivative at $0$ is positive,
then one can strengthen part (i) to say that the unique maximum occurs 
between the means of $h$ and $f$ when $\bal > \frac12$, and
between the means of $h$ and $g$ when $\bal < \frac12$.
In other words, with this assumption, one recovers the additional
qualitative observation that for small separations
between the functions, it is best to give scores that
are slightly above average.
We note that Gaussians are one basic example of a class
of density functions satisfying this condition; there are also
others.  See Figure~\ref{fig:gaussians-synthetic} for an example in which we
plot the mixture when $f$ and $g$ are Gaussian translates, with 
$\bal$ fixed but changing $\controv$ and hence changing the variance.
(Again, it is not necessary to make a Gaussian assumption
for anything we do here; the example is purely for the sake
of concreteness.)

Specifically, our second result is the following.
In its statement, we use $f^{(j)}(x)$ to denote
the $j^{\rm th}$ derivative of a function $f$,
and recall that we say a function is $C^j$ if
it has at least $j$ continuous derivatives.
\begin{theorem}
Suppose we have the hypotheses of Theorem~\ref{thm:general-mixture},
and additionally there is a function $k$ such that
$f(x) = k(x - \mu_f)$ and $g(x) = k(x - \mu_g)$.
(Hence $k$ is unimodal with its unique local maximum at $x = 0$.)

Further, suppose that for some $j$, the function $k$ is $C^{j+1}$ and 
we have $k^{(j)}(0) > 0$ and $k^{(i)}(0) = 0$ for $2 < i < j$.
Then in addition to the conclusions of Theorem~\ref{thm:general-mixture},
we also have 
\begin{itemize}
\item[(i$'$)] 
There exists a constant $\eps_0'$ such that 
when $\controv < \eps_0'$, the combined density 
$h(x)$ has its unique maximum strictly between 
the mean of $f$ and the mean of $h$ when $\bal > \frac12$,
and strictly between the mean of $g$ and the mean of $h$
when $\bal < \frac12$.
\end{itemize}
\label{thm:symmetric-mixture}
\end{theorem}
\xproof{
We omit the proof, which applies Taylor's theorem to $k'$, due to space limitations.
}

We are, of course, not claiming that our model is the only one that
would be consistent with the data we observed; our point is simply to show that there exists at least one simple model that exhibits the desired behavior.

\section{Consistency among countries}
\label{sec:countries}

In this section we evaluate the robustness of the observed
social-effects phenomena by comparing review data from
three additional different national Amazon sites: Amazon.co.uk (U.K),
Amazon.de (Germany) and Amazon.co.jp (Japan), collected
using the same methodology   described
in Section \ref{sec:data},
except that because of the particulars of
  the AWS API, we were unable to filter out mechanically
  cross-posted reviews from the 
Amazon.co.jp data.  It is
reasonable to assume that these reviews were produced
independently by four separate populations of reviewers
(there exist customers who post reviews to multiple Amazon
sites, but such behavior is unusual).

There are noticeable differences between reviews collected
from different regional Amazon sites, in both average
\hratio and review variance (Table \ref{countries}). The
review dynamics in the U.K. and Japan communities appear to
be less controversial than in the U.S. and Germany.
Furthermore, repeating the analysis from
Section \ref{sec:variance} for these three new
datasets reveals the same qualitative patterns observed in
the U.S. data and suggested by the model introduced in
Section \ref{sec:gaussians}.  Curiously enough, for the Japanese data, 
in contrast to its general reputation of a collectivist culture
\cite{Bond+Smith:96a}, 
we observe that 
the left hump
is higher than the right one for reviews with high variance,
 i.e., reviews with
\smarks
\textit{below} the mean are more favored by 
\hevaluators
than the respective reviews with positive deviations (Figure
\ref{fig:countries_var3}).  In the context of our model,
this would correspond to a larger proportion of negative
evaluators (balance 
$p<0.5$). 

\begin{table}
\begin{center}
\begin{tabular}{|c|c|c|c|c|}
\hline
   & Total reviews& Avg h.ratio & Avg \smark var. \\
\hline
U.S. & 1,008,466    & 0.72         & 1.34 \\                
U.K. & 127,195      & 0.80         & 0.95 \\                
Germany & 184,705      & 0.74         & 1.24 \\                
Japan & 253,971      & 0.69         & 0.93 \\                
\hline
\end{tabular} 
\caption{\label{countries} Comparison of review data from four regional sites:
  number of reviews with 10 or more \hvotes, average
  \hratio, and average \variance in \smark.}
\end{center}

\end{table}

\begin{figure}[h]
  \begin{center}
    \begin{minipage}{1.5in}{\includegraphics[width=1.5in]{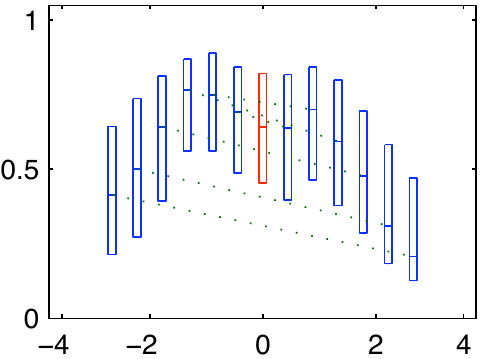}}
    \end{minipage}
    \qquad
    \begin{minipage}{1.5in}{\includegraphics[width=1.5in]{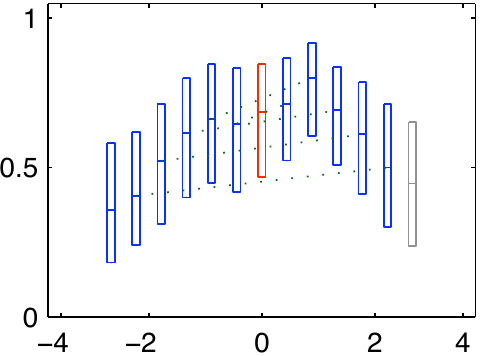}}
    \end{minipage}
    \caption{\label{fig:countries_var3}\Sdeviations vs. \hratio for \variance $=3$, in
      the Japanese (left) and U.S. (right) data. 
      The curve for Japan has a pronounced 
lean towards
the left.}
  \end{center}
\end{figure}

\section{Conclusion}
\label{sec:conclusion}

We have seen that helpfulness evaluations on a site like Amazon.com
provide a way to assess how opinions are evaluated by members
of an on-line community at a very large scale.
A review's 
perceived helpfulness depends not just on its content,
but also the relation of its score to other scores.
This dependence on the score contrasts with a number of
theories from sociology and social psychology, but is
consistent with a simple and natural model of 
individual bias in the presence of a mixture of opinion distributions.

There are a number of interesting directions for further research.
First, the robustness of our results across independent populations
suggests that the phenomenon may be relevant to other settings
in which the evaluation of expressed opinions is a key 
social dynamic.  Moreover, as we have seen in 
Section~\ref{sec:countries}, variations in the effect
(such as the magnitude of deviations above or below the mean)
can be used to form hypotheses about differences in the
collective behaviors of the underlying populations.
Finally, it would also be very interesting to consider 
social feedback mechanisms that might be capable of modifying
the effects we observe here, and to consider the possible
outcomes of such a design problem for systems enabling 
the expression and dissemination of opinions.

\xhdr{Acknowledgments}  
We thank Daria Sorokina, Paul Ginsparg, and Simeon Warner for
assistance with their code, and Michael Macy, 
Trevor Pinch, 
Yongren Shi, 
Felix Weigel, and the anonymous reviewers for
helpful (!)\
comments.
Portions of this work were completed while Gueorgi Kossinets was 
a postdoctoral researcher in
the Department of Sociology at Cornell University.  
This paper is based upon work supported in part by 
a University Fellowship from Cornell, DHS grant N0014-07-1-0152,
 the National Science Foundation 
grants 
 BCS-0537606, 
CCF-0325453, CNS-0403340, and CCF-0728779, 
a John D. and Catherine T. MacArthur Foundation Fellowship, a Google Research
Grant, a Yahoo!\ Research Alliance gift,
a Cornell University Provost's Award for Distinguished Scholarship, a Cornell
University Institute for the Social Sciences Faculty Fellowship, and an Alfred
P.\ Sloan Research Fellowship.  
Any opinions, findings, and conclusions or recommendations expressed are those
of the authors and do not necessarily reflect the views or official policies,
either expressed or implied, of any sponsoring institutions, the U.S.\
government, or any other entity.

\newcommand{\bibsnip}{\vspace{-.09in}}

\end{document}